\documentclass[11pt]{article}

\usepackage[preprint]{acl}

\usepackage{times}
\usepackage{latexsym}
\usepackage{adjustbox}

\usepackage[T1]{fontenc}

\usepackage[utf8]{inputenc}

\usepackage{microtype}
\usepackage{booktabs}
\usepackage{multirow}
\usepackage[table]{xcolor}

\usepackage{inconsolata}

\usepackage{graphicx}

\title{Open, Reliable, and Collective: A Community-Driven Framework for Tool-Using AI Agents}
\author{
 \textbf{Hy Dang\textsuperscript{1}},
 \textbf{Quang Dao\textsuperscript{2}},
 \textbf{Meng Jiang\textsuperscript{1}},
\\
 \textsuperscript{1}University of Notre Dame,
 \textsuperscript{2}Rose-Hulman Institute of Technology,
 \\
 {\includegraphics[height=2ex]{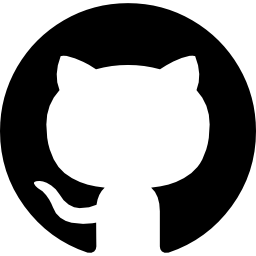}}\,
 \url{https://github.com/hydang99/opentools}
 }

\begin{document}

\maketitle
\begin{abstract}
Tool-integrated LLMs retrieve information, perform computations, and take real-world actions, but their reliability depends on both \emph{tool-use accuracy} and \emph{intrinsic tool accuracy}, including tool correctness, stability, and safety. While prior work primarily emphasizes tool use, intrinsic tool accuracy remains underexamined. We introduce \textbf{\textsc{OpenTools}}\footnote{Source Code and Implementation: \url{https://github.com/hydang99/opentools}}, a community-driven and maintainable toolbox for discovering, using, evaluating, and contributing open-source tools. \textsc{OpenTools} standardizes tool interfaces, converts documented Python functions into reviewable bundles, supports maintainer-triggered evaluation, and combines non-executing risk inspection with optional advisory LLM review. A \textbf{public web demo}\footnote{Web Demonstration: \url{https://huggingface.co/spaces/opentools/opentools}} allows users to run tools and agents, inspect evidence, contribute tests, and submit tools for maintainer review, while MCP enables controlled access from external applications. Experiments show that community-contributed, task-specific tools yield relative gains of 6\% to 22\% over an existing toolbox across multiple agent architectures, highlighting the importance of intrinsic tool accuracy. \textsc{OpenTools} is released under the Apache 2.0 license, with a demonstration video at \url{https://www.youtube.com/watch?v=XhRwATwIBxU}.
\end{abstract}
\section{Introduction}
Large language models (LLMs) have evolved from generative systems into general-purpose agents that can plan, reason, and interact with users over contexts \cite{grattafiori2024llama}. Agentic frameworks support structured reasoning \cite{yao2023tree, li2024chain, zhang2024supervised} and self-reflection \cite{ji2023towards, liu-etal-2025-attention}, extending LLM capabilities beyond question answering and helping mitigate hallucinations \cite{yao2023tree, zhang2024chain, zhu-etal-2025-knowagent}. However, token-only generation remains brittle when tasks require current knowledge or interaction with external systems \cite{paranjape2023art, qin2023toolllm}. Tool-augmented language models (TALMs) address this limitation through retrieval systems \cite{gao2023retrieval, jin2025search}, calculators \cite{schick2023toolformer}, code interpreters \cite{wang2024executable}, and domain APIs \cite{wu2025chemagent, arlt2025towards, jang2025medtutor}. While this delegation improves factuality and supports tasks, it also makes agent reliability dependent on external tools.

We argue that TALM reliability has two failure modes: \textbf{(i) tool-use accuracy}, which concerns selecting and invoking tools correctly, and \textbf{(ii) intrinsic tool accuracy}, which concerns tool correctness, stability, and robustness to drift. Most prior work emphasizes tool learning \cite{qin2024tool, patil2024gorilla} while assuming that tools and their documentation are reliable. In practice, tools can fail because of incomplete coverage, API and dependency changes, nondeterminism, and silent errors. Tool access can also introduce security and credential risks \cite{winston2025taxonomy, milev2025toolfuzz}. Maintaining reliable tools therefore requires a shared interface, representative tests, pre-execution risk inspection, explicit evaluation states, and interoperable access across agent frameworks.

To address these challenges, we introduce \textsc{OpenTools}, an open-source, community-driven toolbox for tool-integrated LLMs. First, \textsc{OpenTools} standardizes tool schemas and wrappers, while a deterministic converter transforms supported Python functions into reviewable bundles with structured metadata. Second, its evaluation and maintenance workflow combines community-contributed tests, non-executing source inspection, secret-risk detection, policy-gated functional evaluation, and optional advisory LLM review of sanitized evidence. Evaluation records distinguish completed, failed, blocked, and not-run states, while continuous integration supports reviewable updates to the shared tool inventory. Third, \textsc{OpenTools} provides a public web interface and a Model Context Protocol (MCP) server. The web interface supports test cases, tool execution, and open-source tool submissions to further improve \textsc{OpenTools}. 

The community remains central to extending tool coverage, contributing tools, failure cases, and reviewing evidence, while automation reduces the effort required to standardize and assess contributions. Across diverse domains, higher-quality, task-specific tool coverage in \textsc{OpenTools} yields \textbf{6\% to 22\%} relative gains over an existing toolbox across multiple agent architectures. We hope \textsc{OpenTools} supports open, reliable, and continuously maintained tool ecosystems for reproducible and trustworthy research.
\section{\textsc{OpenTools} Framework}
\label{opentools}
\begin{figure*}[htbp]
    \centering
    \includegraphics[width=\linewidth]{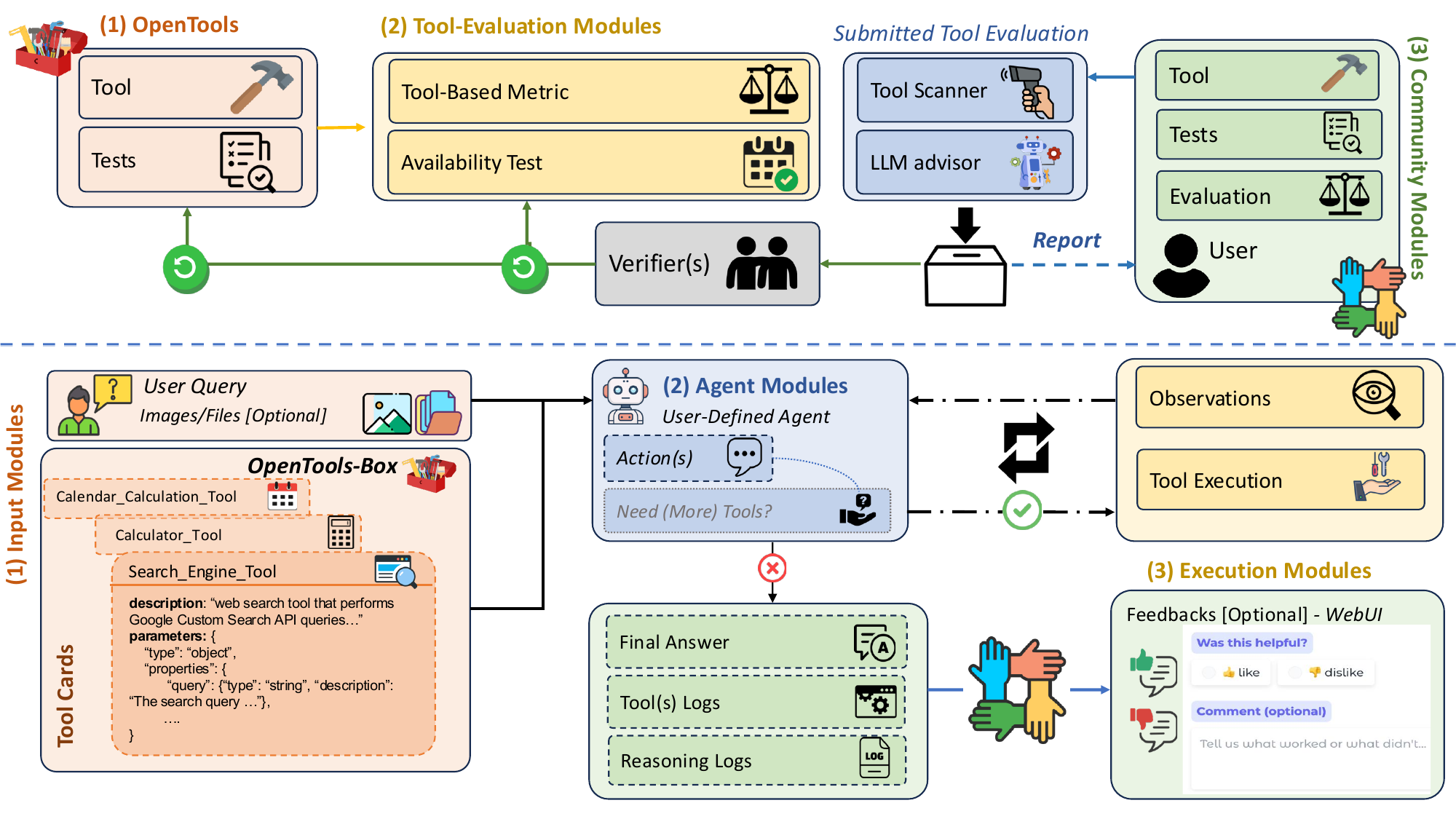}
    \caption{\textsc{OpenTools} supports two complementary workflows. \textbf{Top, tool maintenance:} community members submit tools, tests, prior evaluations, and feedback. Non-executing scanners and an optional LLM advisor produce a review report, while human verifiers make the final acceptance decision and manually update the shared tools, tests, and reliability evidence. Accepted tools are assessed using tool-specific metrics and availability tests. \textbf{Bottom, agentic use:} users provide a query, optional files, and tool cards from the \textsc{OpenTools}-Box. Agent modules select actions and request tool execution, execution modules return observations, and the system produces a final answer with tool and reasoning logs. Users may provide optional feedback through the WebUI.}
    \label{fig:opentool_figure}
\end{figure*}
\textsc{OpenTools} is an open-source, community-driven framework for building, evaluating, sharing, and deploying tools for tool-integrated language models. It supports two complementary workflows, shown in Figure~\ref{fig:opentool_figure}: (1) a \emph{Tool Accuracy and Maintenance Loop} for standardizing contributions, evaluating intrinsic tool reliability, and maintaining the shared toolbox, and (2) an \emph{Agentic Workflow} for integrating tool collections into LLM agents. The framework uses common tool schemas, separates tool maintenance from agent policies, and records structured evaluation and execution evidence for reproducibility and debugging.

\subsection{Tool Accuracy and Maintenance Loop}

An agent may select the correct tool and arguments but still receive an incorrect or unavailable result. Tool behavior can change because of API updates, dependency drift, nondeterminism, rate limits, or backend failures. Workflow~1 therefore treats \emph{intrinsic tool accuracy} as a maintained property rather than a fixed assumption.

\paragraph{Problem Definition.}
Given a tool $\tau$, its source $S_\tau$, a test suite $\mathcal{T}_t$, and an execution policy $P$, \textsc{OpenTools} produces a reliability record
\[
R_t(\tau) =
\texttt{Evaluate}(\tau, S_\tau, \mathcal{T}_t, P).
\]
The record includes the tool version, evaluation, static risk evidence, test outcomes, and execution status. The workflow consists of three stages: contribution and standardization, risk-aware evaluation, and community maintenance.
\subsubsection{Tool Contribution and Standardization}

The key design contribution of \textsc{OpenTools} is a \emph{review-oriented lifecycle} connecting tool standardization with automated evidence and maintainer review. Each tool has a natural-language description, typed JSON schema, output contract, and tool card recording dependencies, provenance, limitations, risk evidence, and evaluation history. For supported Python functions, a \textbf{deterministic converter} derives the schema from type annotations, generates a standardized wrapper, preserves the source, and packages the artifacts for review. Ambiguous entry points and unsupported signatures produce explicit errors rather than inferred conversions.

\subsubsection{Risk-Aware Tool Evaluation}

\paragraph{Layered Non-Executing Inspection.}
Before any permitted functional test, \textsc{OpenTools} inspects submitted source without importing or running it. Its AST analyzer identifies syntax errors, credential access, filesystem writes, network and process capabilities, dynamic execution, and possible embedded secrets. When available, \textsc{Gitleaks} and \textsc{detect-secrets} provide additional secret detection, while \textsc{Bandit} and \textsc{Semgrep} identify Python-specific and rule-based security issues. Their outputs are normalized into reports containing scanner status, rule, severity, source location, and sanitized evidence. Unavailable or failed scanners remain explicit, and potential secret values are redacted. These findings support risk assessment and help maintainers quickly identify potential risks in contributed tools.

\paragraph{Contributor-Facing Advisory Review.}
Contributors receive the generated tool card, conversion status, scanner coverage, detailed findings, and recommended actions. They may also request an optional LLM judge that summarizes concerns, required actions, and a recommendation from sanitized metadata and evidence. The LLM does not receive raw source code or credential values. Instead, contributors can use this evidence to refine their own tool implementations.

\paragraph{Gated Evaluation and Maintainer Review.}
The hosted interface inspects but does not execute newly uploaded tools. Maintainers run functional tests locally for explicitly selected tools that satisfy the configured risk policy. Scanner findings, test results, and optional LLM recommendations provide supporting evidence, while a human maintainer makes the final acceptance decision.

\subsubsection{Community-Driven Maintenance}

Community members contribute tools, tests, and failure reports through the repository or web interface. Maintainers review contribution bundles, and accepted cases become regression tests for later evaluations. To refresh reliability evidence, maintainers can manually reevaluate selected tools, regenerate tool cards and the shared inventory, and review the resulting changes before committing them. This process connects community feedback with repeatable evaluation while preserving maintainer control over the shared toolbox.

\subsection{Agentic Workflow}
Whereas the Maintenance Loop focuses on intrinsic tool reliability, the agentic workflow uses \textsc{OpenTools} as a toolbox for tool-integrated LLM agents. It packages tool collections for general use across agent frameworks and provides end-to-end execution with transparent, debuggable traces.

\paragraph{Problem Definition.}
Given a user query $q$, a toolbox $\mathcal{S}$, and an agent policy $\pi$ (e.g., prompting-based planning/routing or custom control flow), the agentic workflow produces an answer $y$ by executing tool calls and intermediate reasoning steps. It also outputs an execution trace $\mathcal{L}$ that records tool usage and outcomes for reproducibility and failure diagnosis: $(q,\mathcal{S},\pi)\rightarrow(y,\mathcal{L}).$
\paragraph{Workflow Overview.}
The agentic workflow has two phases: \textit{(i) Agent Setup and Tool Access} and \textit{(ii) Interactive Task Execution}.
\subsubsection{Agent Setup and Tool Access}
The setup phase specifies which tools the agent can use and how they are exposed.

\paragraph{Toolbox Interface.}
Agents access the standardized tool representations defined in Workflow~1 through the native \textsc{OpenTools} interface or the Model Context Protocol. MCP supports discovery, inspection, evaluation, and invocation of explicitly allowlisted tools, enabling integration with external applications without tool-specific adapters.

\paragraph{Agent Policy Selection.}
\textsc{OpenTools} includes several standard agent policies (e.g., ReAct \cite{yao2023react}, OctoTools \cite{lu-etal-2026-octotools}, and MultiAgent) and supports user-defined agents via custom controllers and workflows. The policy module determines action selection, and policies are treated as plug-and-play components, decoupling \emph{tool access} from \emph{decision logic}.

\subsubsection{Interactive Task Execution}
During execution, the agent iteratively updates the task state, invokes tools, and produces the final answer. An execution manager mediates between the policy and toolbox: it validates arguments against schemas, executes tool calls, returns results as observations, and records failures (e.g., exceptions, rate limits, timeouts). \textsc{OpenTools} logs a structured trace of tool invocations, inputs/outputs, and error states, enabling reproducibility and attribution of failures to either policy errors (tool-use) or tool-side issues (intrinsic reliability), and linking behavior to Workflow~1 reliability signals. The system outputs the answer and trace for interactive inspection and offline analysis.

\subsection{Web Demonstration Interface}
\label{sec:demonstrationinterface}

\textsc{OpenTools} provides a public web demonstration interface\footnote{\url{https://huggingface.co/spaces/opentools/opentools}} that exposes both workflows without requiring local installation. Users can inspect tool cards and recorded evaluations, run supported tools and predefined agents, contribute test cases, submit open-source tools for review. Additional interface details are provided in Appendix~\ref{sec:web_interface}.

\paragraph{Tool Accuracy and Contribution.}
For Workflow~1, users can inspect schemas and usage guidance, run supported tools on custom inputs, and view the latest recorded evaluation. Individual or bulk test cases can be submitted as candidate regression cases. The \emph{Contribute Tools} interface applies the conversion and non-executing inspection pipeline described above, presents its risk and scanner evidence, and optionally requests advisory LLM review of sanitized evidence. It then creates a pending-review bundle; uploaded programs are neither executed nor automatically published.

\paragraph{Agentic Workflow Interaction.}
For Workflow~2, the demo provides an end-to-end sandbox where users can input a task query, optionally restrict the tool subset, and run a \emph{predefined} agent configuration (e.g., ReAct \cite{yao2023react}, OctoTools \cite{lu-etal-2026-octotools}). The interface returns both the final response and a structured execution trace, including the sequence of tool calls, tool arguments, intermediate observations, and errors if any. Users can provide feedback on both tool-level outputs and the overall agent response, facilitating diagnosis of whether failures arise from tool-use decisions or from intrinsic tool issues.

\section{Experiments}
\begin{table*}[!htbp]
\centering
\small
\setlength{\tabcolsep}{4.2pt}
\renewcommand{\arraystretch}{1}
\begin{tabular}{l c cc cc cc}
\toprule
 & \textbf{Prompting}
 & \multicolumn{2}{c}{\textbf{ReAct}} 
 & \multicolumn{2}{c}{\textbf{OctoTools}} 
 & \multicolumn{2}{c}{\textbf{MultiAgent}} \\
\cmidrule(lr){3-4}\cmidrule(lr){5-6}\cmidrule(lr){7-8}
\textbf{Task} 
& - 
& OctoTools-T & \textit{OpenTools-T}
& OctoTools-T & \textit{OpenTools-T}
& OctoTools-T & \textit{OpenTools-T} \\
\midrule
VQA/Puzzle & 45.38 / 74.36& 55.52 / 73.13 & \textbf{68.95} / \textbf{79.92} & 52.13 / 69.75 & \textbf{67.01} / \textbf{80.48} & 72.40 & \textbf{75.28} \\
Math/Reasoning &  43.77 / 83.58 & 48.71 / 83.67 & \textbf{59.03} / \textbf{85.33} & 42.93 / 78.23 & \textbf{55.54} / \textbf{81.28} & 81.25 & \textbf{83.15} \\
Scientific   &    46.54 / 80.20 & 50.67 / 75.67 & \textbf{53.38} / \textbf{79.89} & 47.87 / 76.11 & \textbf{52.97} / \textbf{77.11} & 69.48 & \textbf{73.78} \\
Medical     &    43.56 / 69.56 & 52.00 / 69.11 & \textbf{57.59} / \textbf{71.52} & 53.00 / 66.78 & \textbf{55.48} / \textbf{68.50} & 66.05 & \textbf{68.67} \\
Agent      &      5.51 / 21.41 & 14.17 / 51.97 & \textbf{28.50} / \textbf{55.91} & 4.72/ 46.03 & \textbf{24.41} /\textbf{53.54} &  53.92 & \textbf{66.14} \\
\midrule
Average    &     42.16 / 73.49  & 49.27 / 74.23 & \textbf{58.22} / \textbf{78.08} & 45.84 / 71.10 & \textbf{56.00} / \textbf{75.83} & 71.67 & \textbf{75.15} \\
\bottomrule
\end{tabular}
\caption{Task-group performance for Prompting and three agent approaches using OctoTools-T vs.\ OpenTools-T. Scores are reported as gpt-4o-mini / gpt-5-mini for Prompting, ReAct, and OctoTools; MultiAgent reports a single gpt-5-mini score. OpenTools-T improves the overall average across all frameworks, demonstrating that higher intrinsic tool quality can translate into better end-to-end reliability. Higher means better.}
\label{tab:opentools_results_summary}
\end{table*}
We evaluate \textsc{OpenTools} under the \emph{agentic workflow} to quantify how a community-maintained toolbox impacts end-to-end task performance across multiple agent frameworks. Concretely, we compare agent performance when equipped with the \textsc{OpenTools} toolbox (including community-contributed tools and tests) versus a strong, curated toolbox baseline (OctoTools toolbox). Our evaluation targets breadth (many benchmarks and domains) and generality (multiple agent policies), while keeping the underlying base LLM fixed.

\subsection{Experimental Setup}
\subsubsection{Benchmarks.}
Following the OctoTools testbed protocol \cite{lu-etal-2026-octotools}, we evaluate a diverse suite of benchmarks spanning multiple domains and reasoning types, grouped into five task categories: \textbf{VQA/Puzzle} (AlgoPuzzleVQA \cite{ghosal2025algopuzzlevqa}, Hallusion-VD \cite{guan2024hallusionbench}, PuzzleVQA \cite{chia2024puzzlevqa}, VQAv2 \cite{goyal2017making}), \textbf{Math/Reasoning} (Game of 24 \cite{lile2024_24game}, Omni-MATH \cite{gao2024omni}, CLEVR-Math \cite{lindstrom2022clevr}, MathVista \cite{lu2023mathvista}), \textbf{Scientific} (GPQA \cite{rein2024gpqa}, MMLU-Pro \cite{wang2024mmlu}, SciFiBench \cite{roberts2024scifibench}), \textbf{Medical} (MedQA \cite{jin2021disease}, PathVQA \cite{he2020pathvqa}, SLAKE \cite{liu2021slake}), and \textbf{Agent} (GAIA-Text \cite{mialon2023gaia}).
\subsubsection{Baselines}
\subsection{Toolboxes}

We compare two experimental toolboxes. \textbf{OctoTools-T} is the 13-tool collection used by the OctoTools testbed \cite{lu-etal-2026-octotools}, including a general solution generator and domain tools for web search, image understanding, code generation, and classification. \textbf{\textsc{OpenTools-T}} contains 42 tools drawn from three sources: general-purpose tools such as search and calculation, curated domain tools for areas such as vision and chemistry, and community-contributed tools such as specialized maze solvers. All \textsc{OpenTools-T} tools follow the common interface described in Section~\ref{opentools} and can be reevaluated through the Tool Accuracy and Maintenance Loop.

We categorize tools by their primary execution mechanism. \textbf{Program-based} tools perform local computation without relying on remote services. \textbf{API-based} tools wrap externally hosted services, which may require user-provided credentials. \textbf{Prompting-based} tools implement model-mediated functions through fixed prompt templates. Table~\ref{tab:toolbox_stats} summarizes the resulting composition.

\begin{table}[t]
\centering
\small
\begin{tabular}{lrr}
\toprule
 & OctoTools-T & \textsc{OpenTools-T} \\
\midrule
Total tools      & 13 & 42 \\
Program-based    & 1  & 16 \\
API-based        & 8  & 19 \\
Prompting-based  & 4  & 7  \\
\bottomrule
\end{tabular}
\caption{Composition of the evaluated toolboxes by primary execution mechanism.}
\label{tab:toolbox_stats}
\end{table}

The \textsc{OpenTools-T} implementations include newly developed tools and open-source wrappers adapted from ecosystems such as OctoTools \cite{lu-etal-2026-octotools} and AWorld \cite{yu2025aworld}. Adapted tools are refactored to use consistent schemas across agent policies. This broader composition allows us to examine how tool coverage and intrinsic tool quality affect end-to-end agent performance.
\paragraph{Agent Frameworks.}
We evaluate our systems under multiple agent frameworks: \textbf{(i) Prompting:} the LLM answers directly without tool use. We experiment with both zero-shot and chain of thought prompting (CoT), which are step-by-step reasoning without external tools). In preliminary experiments, zero-shot performs better overall for both gpt-4o-mini and gpt-5-mini; therefore, we report zero-shot as \emph{Prompting} in the main results and provide the full zero-shot/CoT comparison in Appendix~\ref{sec:experiment_result}. \textbf{(ii) ReAct Agent \cite{yao2023react}:} an iterative reason-and-act agent that interleaves natural-language reasoning with tool calls. \textbf{(iii) OctoTools Agent \cite{lu-etal-2026-octotools}:} the planner--executor style agent as defined in the OctoTools framework. \textbf{(iv) MultiAgent Framework:} Inspired by LemonAgent \cite{jiang2026lemon}, we use a hierarchical pipeline that decomposes a query and routes sub-problems to specialized sub-agents. Unlike LemonAgent, we (i) perform explicit reasoning-based decomposition before routing and (ii) use a \emph{Verifier}/guardrail to store only validated intermediate results in shared memory. A \emph{Planner} decomposes the query, a \emph{Generator} issues tool calls, tools are executed and verified, validated results are written to memory, and a \emph{Composer} generates the final answer with a brief memory-based summary.

For tool-using agents, we run each framework with (i) the OctoTools toolbox (OctoTools-T) and (ii) our \textsc{OpenTools} toolbox (OpenTools-T), enabling a controlled comparison of toolsets under the same decision logic. Following OctoTools~\cite{lu-etal-2026-octotools}, we perform greedy tool selection for OpenTools-T to choose task-appropriate tool subsets. For OctoTools-T, we use the tool sets reported in the OctoTools paper that achieve the best performance for each task.

\section{Results \& Analysis}
Table~\ref{tab:opentools_results_summary} compares OctoTools (OctoTools-T) and our OpenTools toolbox (\textsc{OpenTools-T}) under agent frameworks. We report task-group accuracy in Table~\ref{tab:opentools_results_summary} using the same sampled instance identifiers as OctoTools for direct comparability. We run three independent trials per setting and report per-benchmark mean $\pm$ std.\ in Appendix~\ref{sec:appendix}. 

\paragraph{Tool Augmentation Effect.}
For the weaker model (gpt-4o-mini), tool augmentation consistently outperforms tool-free prompting. With OctoTools-T, ReAct and OctoTools gain \textbf{+16.86\%} and \textbf{+8.70\%} (relative) over Prompting; with \textsc{OpenTools-T}, the gains rise to \textbf{+38.09\%} and \textbf{+32.83\%}. This indicates that tools substantially boost smaller LLMs. For the stronger model (gpt-5-mini), tool-free prompting is already strong on knowledge/reasoning-heavy tasks (VQA/Puzzle, Math/Reasoning, Scientific, Medical), so tool gains are smaller and can even turn negative with weaker toolbox. With OctoTools-T, \textbf{ReAct} improves over Prompting by only \textbf{+1\%} on average, and \textbf{OctoTools} can underperform Prompting. In contrast, with \textsc{OpenTools-T}, both \textbf{ReAct} and \textbf{OctoTools} outperform \textbf{Prompting} (78.08 and 75.83 vs.\ 73.49 on average), indicating that \emph{tool reliability and coverage} still matter even for strong base models.

Tools matter most on the hardest Agent task, where external actions are required. Compared to \textbf{Prompting}, \textbf{ReAct} with OctoTools-T/\textsc{OpenTools-T} improves by \textbf{143\%}/\textbf{161\%} (relative), and \textbf{OctoTools} with OctoTools-T/\textsc{OpenTools-T} improves by \textbf{115\%}/\textbf{150\%}.
\paragraph{Better Tools Yield Consistent Gains.}
Replacing OctoTools-T with \textsc{OpenTools-T} improves performance for every agent framework. On average, \textsc{OpenTools-T} yields relative gains of \textbf{18.16\%/5.19\%} for \textbf{ReAct} and \textbf{22.16\%/6.7\%} for \textbf{OctoTools} (gpt-4o-mini / gpt-5-mini), and improves \textbf{MultiAgent} by \textbf{4.86\%}. These gains are consistent across domains, indicating that a more comprehensive and reliable toolbox translates into better end-to-end agent performance.

The largest improvements concentrate on the \textbf{Agent} task, where tool breadth and robustness are crucial. With the stronger model, \textsc{OpenTools-T} improves over OctoTools-T by \textbf{+7.58\%} (\textbf{ReAct}), \textbf{+16.32\%} (\textbf{OctoTools}), and \textbf{+22.67\%} (\textbf{MultiAgent}) on Agent. \textbf{MultiAgent} with \textsc{OpenTools-T} achieves the best Agent score overall (66.14), but we also observe that complex agent policies can hurt on simpler tasks when paired with a less reliable toolbox (e.g., \textbf{MultiAgent} with OctoTools-T underperforms \textbf{Prompting} on average). Overall, \textsc{OpenTools-T} provides the most consistent gains, especially for tool-intensive end-goal tasks.

\section{Related Work}
\paragraph{Tool use and agentic LLMs.}
Prior work on tool use and agentic LLMs focuses on how models plan, and invoke external tools through training or prompting, including Toolformer-style learning \cite{schick2023toolformer}, ReAct reasoning-and-acting \cite{yao2023react}, and agent controllers that add reflection/search and multi-step tool execution \cite{patil2024gorilla, dang-etal-2025-improving}. These lines of work primarily improve tool-use accuracy, while often assuming the underlying tools are correct and stable.

In parallel, tool libraries and evaluations (e.g., LangChain\footnote{https://github.com/langchain-ai/langchain}, AutoGen\footnote{https://github.com/microsoft/autogen}) broaden tool access and provide domain-specific suites (e.g., chemistry \cite{wu2025chemagent}, medical \cite{li2024mmedagent}, vision \cite{hu2024visual}). Existing benchmarks assess end-to-end task success and tool-calling competence \cite{huang2023metatool, guo2024stabletoolbench}, but frequently conflate model mistakes with tool-side failures and seldom measure tool drift or regressions over time. \textsc{OpenTools} addresses this infrastructure gap by standardizing JSON-schema tool interfaces, providing community-maintained wrappers, and enabling intrinsic tool evaluation with reliability signals and regression tracking—supporting plug-and-play use across agent frameworks while separating tool-use from tool reliability.

\section{Conclusion and Future Work}

We introduced \textsc{OpenTools}, a community-driven and maintainable toolbox for discovering, contributing, evaluating, and using open-source tools with LLM agents. The framework distinguishes \emph{tool-use accuracy} from \emph{intrinsic tool accuracy} and provides a review-oriented lifecycle combining standardized tool interfaces, automated risk and reliability evidence, and maintainer verification. Its public web interface supports community participation, while MCP enables controlled integration with external applications. Experiments show that higher-quality, task-specific tools yield consistent improvements across agent policies, including relative gains of 6\% to 22\% over an existing toolbox. 
\paragraph{Future work} will expand domain-specific tools and regression tests, strengthen isolated and longitudinal evaluation, and study community maintenance at scale. Although layered scanning and advisory LLM review provide useful evidence, they cannot guarantee tool correctness or safety. A well-defined and accountable contribution process, with human oversight of acceptance and publication, will therefore remain central to \textsc{OpenTools}.
\bibliography{main}

@article{gao2023retrieval,
  title={Retrieval-augmented generation for large language models: A survey},
  author={Gao, Yunfan and Xiong, Yun and Gao, Xinyu and Jia, Kangxiang and Pan, Jinliu and Bi, Yuxi and Dai, Yi and Sun, Jiawei and Wang, Haofen and Wang, Haofen},
  journal={arXiv preprint arXiv:2312.10997},
  volume={2},
  year={2023}
}

@inproceedings{yao2023react,
  title={React: Synergizing reasoning and acting in language models},
  author={Yao, Shunyu and Zhao, Jeffrey and Yu, Dian and Du, Nan and Shafran, Izhak and Narasimhan, Karthik and Cao, Yuan},
  booktitle={International Conference on Learning Representations (ICLR)},
  year={2023}
}

@inproceedings{guo2024stabletoolbench,
    title = "{S}table{T}ool{B}ench: Towards Stable Large-Scale Benchmarking on Tool Learning of Large Language Models",
    author = "Guo, Zhicheng  and
      Cheng, Sijie  and
      Wang, Hao  and
      Liang, Shihao  and
      Qin, Yujia  and
      Li, Peng  and
      Liu, Zhiyuan  and
      Sun, Maosong  and
      Liu, Yang",
    editor = "Ku, Lun-Wei  and
      Martins, Andre  and
      Srikumar, Vivek",
    booktitle = "Findings of the Association for Computational Linguistics: ACL 2024",
    month = aug,
    year = "2024",
    address = "Bangkok, Thailand",
    publisher = "Association for Computational Linguistics",
    url = "https://aclanthology.org/2024.findings-acl.664/",
    doi = "10.18653/v1/2024.findings-acl.664",
    pages = "11143--11156"
}

@article{paranjape2023art,
  title={Art: Automatic multi-step reasoning and tool-use for large language models},
  author={Paranjape, Bhargavi and Lundberg, Scott and Singh, Sameer and Hajishirzi, Hannaneh and Zettlemoyer, Luke and Ribeiro, Marco Tulio},
  journal={arXiv preprint arXiv:2303.09014},
  year={2023}
}

@article{patil2024gorilla,
  title={Gorilla: Large language model connected with massive apis},
  author={Patil, Shishir G and Zhang, Tianjun and Wang, Xin and Gonzalez, Joseph E},
  journal={Advances in Neural Information Processing Systems},
  volume={37},
  pages={126544--126565},
  year={2024}
}

@article{schick2023toolformer,
  title={Toolformer: Language models can teach themselves to use tools},
  author={Schick, Timo and Dwivedi-Yu, Jane and Dess{\`\i}, Roberto and Raileanu, Roberta and Lomeli, Maria and Hambro, Eric and Zettlemoyer, Luke and Cancedda, Nicola and Scialom, Thomas},
  journal={Advances in Neural Information Processing Systems},
  volume={36},
  pages={68539--68551},
  year={2023}
}

@article{qin2023toolllm,
  author       = {Yujia Qin and
                  Shihao Liang and
                  Yining Ye and
                  Kunlun Zhu and
                  Lan Yan and
                  Yaxi Lu and
                  Yankai Lin and
                  Xin Cong and
                  Xiangru Tang and
                  Bill Qian and
                  Sihan Zhao and
                  Runchu Tian and
                  Ruobing Xie and
                  Jie Zhou and
                  Mark Gerstein and
                  Dahai Li and
                  Zhiyuan Liu and
                  Maosong Sun},
  title        = {ToolLLM: Facilitating Large Language Models to Master 16000+ Real-world
                  APIs},
  journal      = {CoRR},
  volume       = {abs/2307.16789},
  year         = {2023},
  url          = {https://doi.org/10.48550/arXiv.2307.16789},
  doi          = {10.48550/ARXIV.2307.16789},
  eprinttype    = {arXiv},
  eprint       = {2307.16789},
  bibsource    = {dblp computer science bibliography, https://dblp.org}
}

@article{zhang2024supervised,
  title={Supervised chain of thought},
  author={Zhang, Xiang and Ding, Dujian},
  journal={arXiv preprint arXiv:2410.14198},
  year={2024}
}

@article{li2024chain,
  title={Chain of thought empowers transformers to solve inherently serial problems},
  author={Li, Zhiyuan and Liu, Hong and Zhou, Denny and Ma, Tengyu},
  journal={arXiv preprint arXiv:2402.12875},
  volume={1},
  year={2024}
}

@article{huang2023metatool,
  title   = {MetaTool Benchmark: Deciding Whether to Use Tools and Which to Use},
  author  = {Yue Huang and Jiawen Shi and Yuan Li and Chenrui Fan and Siyuan Wu and Qihui Zhang and Yixin Liu and Pan Zhou and Yao Wan and Neil Zhenqiang Gong and Lichao Sun},
  year    = {2023},
  journal = {arXiv preprint arXiv: 2310.03128}
}

@article{grattafiori2024llama,
  title={The llama 3 herd of models},
  author={Grattafiori, Aaron and Dubey, Abhimanyu and Jauhri, Abhinav and Pandey, Abhinav and Kadian, Abhishek and Al-Dahle, Ahmad and Letman, Aiesha and Mathur, Akhil and Schelten, Alan and Vaughan, Alex and others},
  journal={arXiv preprint arXiv:2407.21783},
  year={2024}
}

@inproceedings{zhu-etal-2025-knowagent,
    title = "{K}now{A}gent: Knowledge-Augmented Planning for {LLM}-Based Agents",
    author = "Zhu, Yuqi  and
      Qiao, Shuofei  and
      Ou, Yixin  and
      Deng, Shumin  and
      Lyu, Shiwei  and
      Shen, Yue  and
      Liang, Lei  and
      Gu, Jinjie  and
      Chen, Huajun  and
      Zhang, Ningyu",
    editor = "Chiruzzo, Luis  and
      Ritter, Alan  and
      Wang, Lu",
    booktitle = "Findings of the Association for Computational Linguistics: NAACL 2025",
    month = apr,
    year = "2025",
    address = "Albuquerque, New Mexico",
    publisher = "Association for Computational Linguistics",
    url = "https://aclanthology.org/2025.findings-naacl.205/",
    pages = "3709--3732",
    ISBN = "979-8-89176-195-7"
}

@article{qin2024tool,
  title={Tool learning with foundation models},
  author={Qin, Yujia and Hu, Shengding and Lin, Yankai and Chen, Weize and Ding, Ning and Cui, Ganqu and Zeng, Zheni and Zhou, Xuanhe and Huang, Yufei and Xiao, Chaojun and others},
  journal={ACM Computing Surveys},
  volume={57},
  number={4},
  pages={1--40},
  year={2024},
  publisher={ACM New York, NY}
}

@article{zhang2024chain,
  title={Chain of preference optimization: Improving chain-of-thought reasoning in llms},
  author={Zhang, Xuan and Du, Chao and Pang, Tianyu and Liu, Qian and Gao, Wei and Lin, Min},
  journal={Advances in Neural Information Processing Systems},
  volume={37},
  pages={333--356},
  year={2024}
}

@article{yao2023tree,
  title={Tree of thoughts: Deliberate problem solving with large language models},
  author={Yao, Shunyu and Yu, Dian and Zhao, Jeffrey and Shafran, Izhak and Griffiths, Tom and Cao, Yuan and Narasimhan, Karthik},
  journal={Advances in neural information processing systems},
  volume={36},
  pages={11809--11822},
  year={2023}
}

@inproceedings{ji2023towards,
  title={Towards mitigating LLM hallucination via self reflection},
  author={Ji, Ziwei and Yu, Tiezheng and Xu, Yan and Lee, Nayeon and Ishii, Etsuko and Fung, Pascale},
  booktitle={Findings of the Association for Computational Linguistics: EMNLP 2023},
  pages={1827--1843},
  year={2023}
}

@inproceedings{liu-etal-2025-attention,
    title = "Attention-guided Self-reflection for Zero-shot Hallucination Detection in Large Language Models",
    author = "Liu, Qiang  and
      Chen, Xinlong  and
      Ding, Yue  and
      Song, Bowen  and
      Wang, Weiqiang  and
      Wu, Shu  and
      Wang, Liang",
    editor = "Christodoulopoulos, Christos  and
      Chakraborty, Tanmoy  and
      Rose, Carolyn  and
      Peng, Violet",
    booktitle = "Proceedings of the 2025 Conference on Empirical Methods in Natural Language Processing",
    month = nov,
    year = "2025",
    address = "Suzhou, China",
    publisher = "Association for Computational Linguistics",
    url = "https://aclanthology.org/2025.emnlp-main.1063/",
    doi = "10.18653/v1/2025.emnlp-main.1063",
    pages = "21016--21032",
    ISBN = "979-8-89176-332-6"
}

@article{jin2025search,
  title={Search-r1: Training llms to reason and leverage search engines with reinforcement learning},
  author={Jin, Bowen and Zeng, Hansi and Yue, Zhenrui and Yoon, Jinsung and Arik, Sercan and Wang, Dong and Zamani, Hamed and Han, Jiawei},
  journal={arXiv preprint arXiv:2503.09516},
  year={2025}
}

@inproceedings{wang2024executable,
  title={Executable code actions elicit better llm agents},
  author={Wang, Xingyao and Chen, Yangyi and Yuan, Lifan and Zhang, Yizhe and Li, Yunzhu and Peng, Hao and Ji, Heng},
  booktitle={Forty-first International Conference on Machine Learning},
  year={2024}
}

@article{wu2025chemagent,
  title={ChemAgent: Enhancing LLMs for Chemistry and Materials Science through Tree-Search Based Tool Learning},
  author={Wu, Mengsong and Wang, YaFei and Ming, Yidong and An, Yuqi and Wan, Yuwei and Chen, Wenliang and Lin, Binbin and Li, Yuqiang and Xie, Tong and Zhou, Dongzhan},
  journal={arXiv preprint arXiv:2506.07551},
  year={2025}
}

@article{arlt2025towards,
  title={Towards autonomous quantum physics research using LLM agents with access to intelligent tools},
  author={Arlt, S{\"o}ren and Gu, Xuemei and Krenn, Mario},
  journal={arXiv preprint arXiv:2511.11752},
  year={2025}
}

@inproceedings{jang2025medtutor,
  title={MedTutor: A Retrieval-Augmented LLM System for Case-Based Medical Education},
  author={Jang, Dongsuk and Shangguan, Ziyao and Tegtmeyer, Kyle and Gupta, Anurag and Czerminski, Jan T and Chheang, Sophie and Cohan, Arman},
  booktitle={Proceedings of the 2025 Conference on Empirical Methods in Natural Language Processing: System Demonstrations},
  pages={319--353},
  year={2025}
}

@inproceedings{dang-etal-2025-improving,
    title = "Improving Large Language Models Function Calling and Interpretability via Guided-Structured Templates",
    author = "Dang, Hy  and
      Liu, Tianyi  and
      Wu, Zhuofeng  and
      Yang, Jingfeng  and
      Jiang, Haoming  and
      Yang, Tao  and
      Chen, Pei  and
      Wang, Zhengyang  and
      Wang, Helen  and
      Li, Huasheng  and
      Yin, Bing  and
      Jiang, Meng",
    editor = "Christodoulopoulos, Christos  and
      Chakraborty, Tanmoy  and
      Rose, Carolyn  and
      Peng, Violet",
    booktitle = "Proceedings of the 2025 Conference on Empirical Methods in Natural Language Processing",
    month = nov,
    year = "2025",
    address = "Suzhou, China",
    publisher = "Association for Computational Linguistics",
    url = "https://aclanthology.org/2025.emnlp-main.1242/",
    doi = "10.18653/v1/2025.emnlp-main.1242",
    pages = "24437--24453",
    ISBN = "979-8-89176-332-6"
}

@inproceedings{winston2025taxonomy,
  title={A taxonomy of failures in tool-augmented llms},
  author={Winston, Cailin and Just, Ren{\'e}},
  booktitle={2025 IEEE/ACM International Conference on Automation of Software Test (AST)},
  pages={125--135},
  year={2025},
  organization={IEEE}
}

@article{milev2025toolfuzz,
  title={ToolFuzz--Automated Agent Tool Testing},
  author={Milev, Ivan and Balunovi{\'c}, Mislav and Baader, Maximilian and Vechev, Martin},
  journal={arXiv preprint arXiv:2503.04479},
  year={2025}
}

@inproceedings{lu-etal-2026-octotools,
    title = "{O}cto{T}ools: A Multi-Agent Framework with Extensible Tools for Complex Reasoning",
    author = "Lu, Pan  and
      Chen, Bowen  and
      Liu, Sheng  and
      Thapa, Rahul  and
      Boen, Joseph  and
      Zou, James",
    editor = "Liakata, Maria  and
      Moreira, Viviane P.  and
      Zhang, Jiajun  and
      Jurgens, David",
    booktitle = "Proceedings of the 64th Annual Meeting of the {A}ssociation for {C}omputational {L}inguistics (Volume 1: Long Papers)",
    month = jul,
    year = "2026",
    address = "San Diego, California, United States",
    publisher = "Association for Computational Linguistics",
    url = "https://aclanthology.org/2026.acl-long.1/",
    doi = "10.18653/v1/2026.acl-long.1",
    pages = "1--86",
    ISBN = "979-8-89176-390-6"
}

@inproceedings{ghosal2025algopuzzlevqa,
  title={ALGOPUZZLEVQA: Diagnosing Multimodal Reasoning Challenges of Language Models with Algorithmic Multimodal Puzzles},
  author={Ghosal, Deepanway and Toh, Vernon and Chia, Yew Ken and Poria, Soujanya},
  booktitle={Proceedings of the 2025 Conference of the Nations of the Americas Chapter of the Association for Computational Linguistics: Human Language Technologies (Volume 1: Long Papers)},
  pages={9615--9632},
  year={2025}
}

@inproceedings{guan2024hallusionbench,
  title={Hallusionbench: an advanced diagnostic suite for entangled language hallucination and visual illusion in large vision-language models},
  author={Guan, Tianrui and Liu, Fuxiao and Wu, Xiyang and Xian, Ruiqi and Li, Zongxia and Liu, Xiaoyu and Wang, Xijun and Chen, Lichang and Huang, Furong and Yacoob, Yaser and others},
  booktitle={Proceedings of the IEEE/CVF Conference on Computer Vision and Pattern Recognition},
  pages={14375--14385},
  year={2024}
}

@inproceedings{chia2024puzzlevqa,
  title={Puzzlevqa: Diagnosing multimodal reasoning challenges of language models with abstract visual patterns},
  author={Chia, Yew Ken and Toh, Vernon and Ghosal, Deepanway and Bing, Lidong and Poria, Soujanya},
  booktitle={Findings of the Association for Computational Linguistics: ACL 2024},
  pages={16259--16273},
  year={2024}
}

@inproceedings{goyal2017making,
  title={Making the v in vqa matter: Elevating the role of image understanding in visual question answering},
  author={Goyal, Yash and Khot, Tejas and Summers-Stay, Douglas and Batra, Dhruv and Parikh, Devi},
  booktitle={Proceedings of the IEEE conference on computer vision and pattern recognition},
  pages={6904--6913},
  year={2017}
}

@article{gao2024omni,
  title={Omni-math: A universal olympiad level mathematic benchmark for large language models},
  author={Gao, Bofei and Song, Feifan and Yang, Zhe and Cai, Zefan and Miao, Yibo and Dong, Qingxiu and Li, Lei and Ma, Chenghao and Chen, Liang and Xu, Runxin and others},
  journal={arXiv preprint arXiv:2410.07985},
  year={2024}
}

@article{lindstrom2022clevr,
  title={Clevr-math: A dataset for compositional language, visual and mathematical reasoning},
  author={Lindstr{\"o}m, Adam Dahlgren and Abraham, Savitha Sam},
  journal={arXiv preprint arXiv:2208.05358},
  year={2022}
}

@article{lu2023mathvista,
  title={Mathvista: Evaluating mathematical reasoning of foundation models in visual contexts},
  author={Lu, Pan and Bansal, Hritik and Xia, Tony and Liu, Jiacheng and Li, Chunyuan and Hajishirzi, Hannaneh and Cheng, Hao and Chang, Kai-Wei and Galley, Michel and Gao, Jianfeng},
  journal={arXiv preprint arXiv:2310.02255},
  year={2023}
}

@inproceedings{rein2024gpqa,
  title={Gpqa: A graduate-level google-proof q\&a benchmark},
  author={Rein, David and Hou, Betty Li and Stickland, Asa Cooper and Petty, Jackson and Pang, Richard Yuanzhe and Dirani, Julien and Michael, Julian and Bowman, Samuel R},
  booktitle={First Conference on Language Modeling},
  year={2024}
}

@article{wang2024mmlu,
  title={Mmlu-pro: A more robust and challenging multi-task language understanding benchmark},
  author={Wang, Yubo and Ma, Xueguang and Zhang, Ge and Ni, Yuansheng and Chandra, Abhranil and Guo, Shiguang and Ren, Weiming and Arulraj, Aaran and He, Xuan and Jiang, Ziyan and others},
  journal={Advances in Neural Information Processing Systems},
  volume={37},
  pages={95266--95290},
  year={2024}
}

@article{roberts2024scifibench,
  title={Scifibench: Benchmarking large multimodal models for scientific figure interpretation},
  author={Roberts, Jonathan and Han, Kai and Houlsby, Neil and Albanie, Samuel},
  journal={Advances in Neural Information Processing Systems},
  volume={37},
  pages={18695--18728},
  year={2024}
}

@inproceedings{mialon2023gaia,
  title={Gaia: a benchmark for general ai assistants},
  author={Mialon, Gr{\'e}goire and Fourrier, Cl{\'e}mentine and Wolf, Thomas and LeCun, Yann and Scialom, Thomas},
  booktitle={The Twelfth International Conference on Learning Representations},
  year={2023}
}

@inproceedings{liu2021slake,
  title={Slake: A semantically-labeled knowledge-enhanced dataset for medical visual question answering},
  author={Liu, Bo and Zhan, Li-Ming and Xu, Li and Ma, Lin and Yang, Yan and Wu, Xiao-Ming},
  booktitle={2021 IEEE 18th international symposium on biomedical imaging (ISBI)},
  pages={1650--1654},
  year={2021},
  organization={IEEE}
}

@article{he2020pathvqa,
  title={Pathvqa: 30000+ questions for medical visual question answering},
  author={He, Xuehai and Zhang, Yichen and Mou, Luntian and Xing, Eric and Xie, Pengtao},
  journal={arXiv preprint arXiv:2003.10286},
  year={2020}
}

@inproceedings{li2024mmedagent,
  title={Mmedagent: Learning to use medical tools with multi-modal agent},
  author={Li, Binxu and Yan, Tiankai and Pan, Yuanting and Luo, Jie and Ji, Ruiyang and Ding, Jiayuan and Xu, Zhe and Liu, Shilong and Dong, Haoyu and Lin, Zihao and others},
  booktitle={Findings of the Association for Computational Linguistics: EMNLP 2024},
  pages={8745--8760},
  year={2024}
}

@article{hu2024visual,
  title={Visual sketchpad: Sketching as a visual chain of thought for multimodal language models},
  author={Hu, Yushi and Shi, Weijia and Fu, Xingyu and Roth, Dan and Ostendorf, Mari and Zettlemoyer, Luke and Smith, Noah A and Krishna, Ranjay},
  journal={Advances in Neural Information Processing Systems},
  volume={37},
  pages={139348--139379},
  year={2024}
}

@article{jin2021disease,
  title={What disease does this patient have? a large-scale open domain question answering dataset from medical exams},
  author={Jin, Di and Pan, Eileen and Oufattole, Nassim and Weng, Wei-Hung and Fang, Hanyi and Szolovits, Peter},
  journal={Applied Sciences},
  volume={11},
  number={14},
  pages={6421},
  year={2021},
  publisher={MDPI}
}

@misc{lile2024_24game,
  author       = {Lile, Nathan},
  title        = {Math Twenty Four (24s Game) Dataset},
  year         = {2024},
  howpublished = {\url{https://huggingface.co/datasets/nlile/24-game}},
  note         = {Hugging Face Datasets}
}

@article{yu2025aworld,
  title={Aworld: Orchestrating the training recipe for agentic ai},
  author={Yu, Chengyue and Lu, Siyuan and Zhuang, Chenyi and Wang, Dong and Wu, Qintong and Li, Zongyue and Gan, Runsheng and Wang, Chunfeng and Hou, Siqi and Huang, Gaochi and others},
  journal={arXiv preprint arXiv:2508.20404},
  year={2025}
}

@article{jiang2026lemon,
  title={Lemon Agent Technical Report},
  author={Jiang, Haipeng and Ren, Kailong and Yin, Zimo and Sun, Zhetao and Gan, Xin and Lv, Guangyi and He, Ming and Wang, Peng and Yin, Congli and Pan, Hong and others},
  journal={arXiv preprint arXiv:2602.07092},
  year={2026}
}
\appendix
\clearpage
\section{Appendix}
\label{sec:appendix}
\subsection{Web Demonstration Interface}
\label{sec:web_interface}
\begin{figure}[h]
    \centering
    \includegraphics[width=\textwidth]{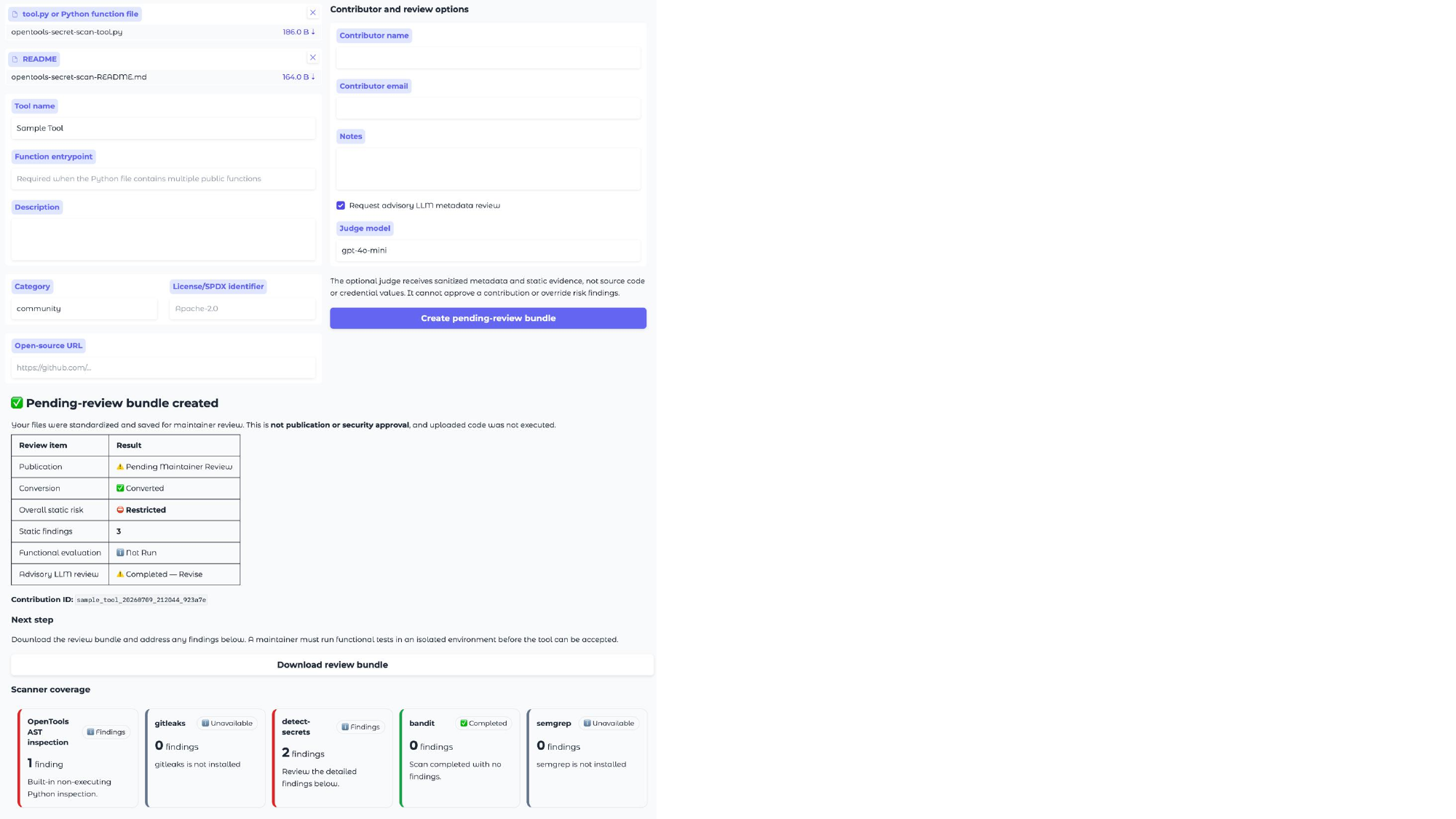}
    \caption{\textsc{OpenTools} tool-contribution interface. Contributors upload a Python implementation and README, provide metadata, and may request advisory LLM review. The interface performs deterministic conversion and non-executing inspection, then reports conversion status, static risk, scanner coverage and findings, and the LLM recommendation. Uploaded code remains unexecuted and unpublished, and the resulting pending-review bundle is returned for contributor revision and maintainer review.}
    \label{fig:tool_contribution}
\end{figure}

\paragraph{Tool Contribution.}
As described in Section~\ref{sec:demonstrationinterface}, users can contribute an open-source tool by uploading Python and README files, providing tool metadata, and optionally requesting LLM review. The interface applies deterministic conversion and non-executing inspection, then returns a pending-review bundle containing the standardized wrapper, tool card, and review evidence. The contributor-facing report presents conversion status, overall risk, scanner coverage, detailed findings, recommended actions, and the optional LLM recommendation. Unavailable scanners are reported explicitly rather than treated as successful checks. Uploaded code is not executed or automatically published; maintainers review the bundle and run functional tests separately before acceptance. Figure~\ref{fig:tool_contribution} shows the submission and reporting interfaces.

\begin{figure}
    \centering
    \includegraphics[width=\linewidth]{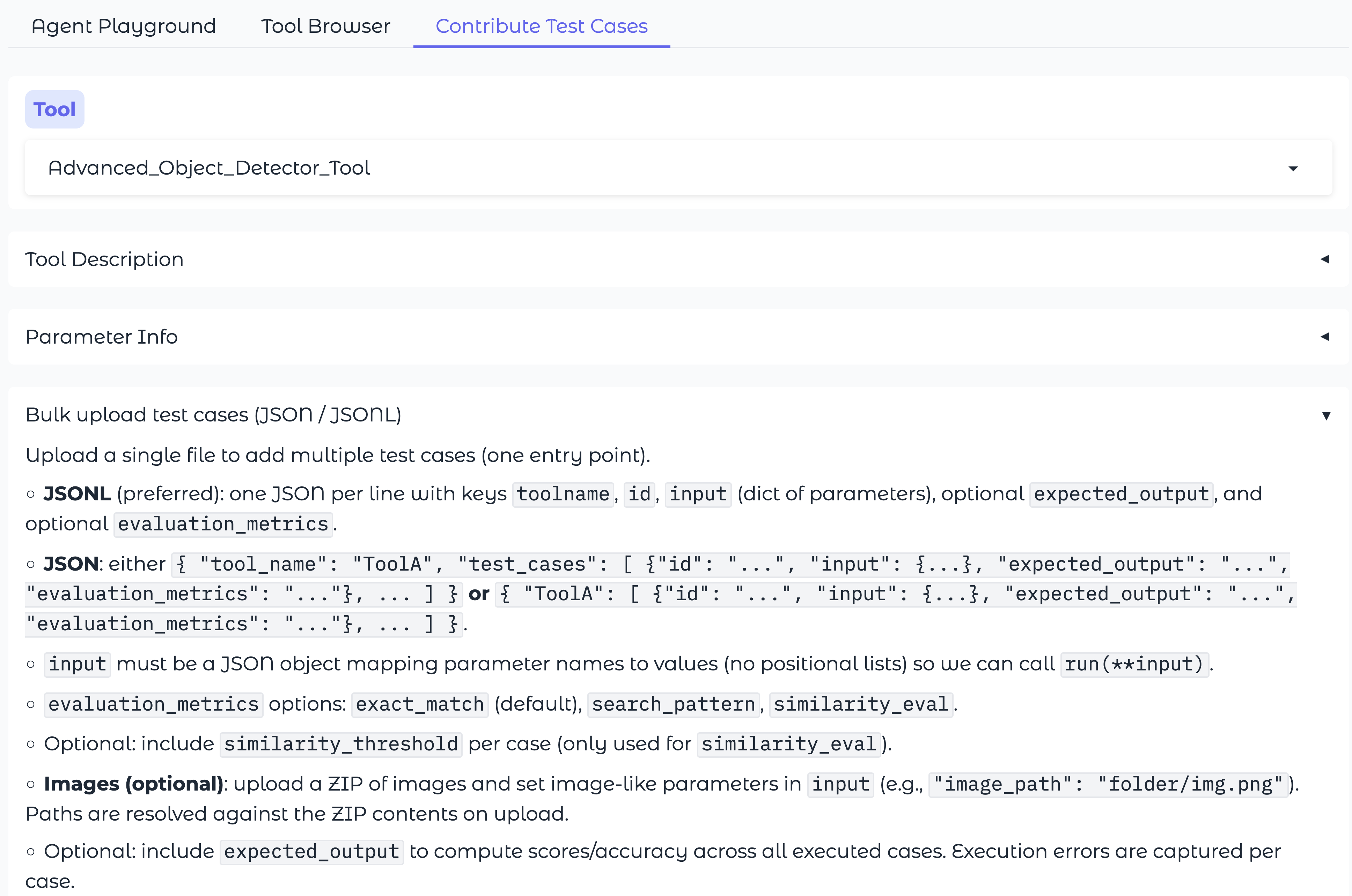}
    \caption{Test-case contribution UI: submission instructions and supported formats for adding new tool tests.}
    \label{fig:instruct}
\end{figure}

\begin{figure}[h]
    \centering
    \includegraphics[width=\linewidth]{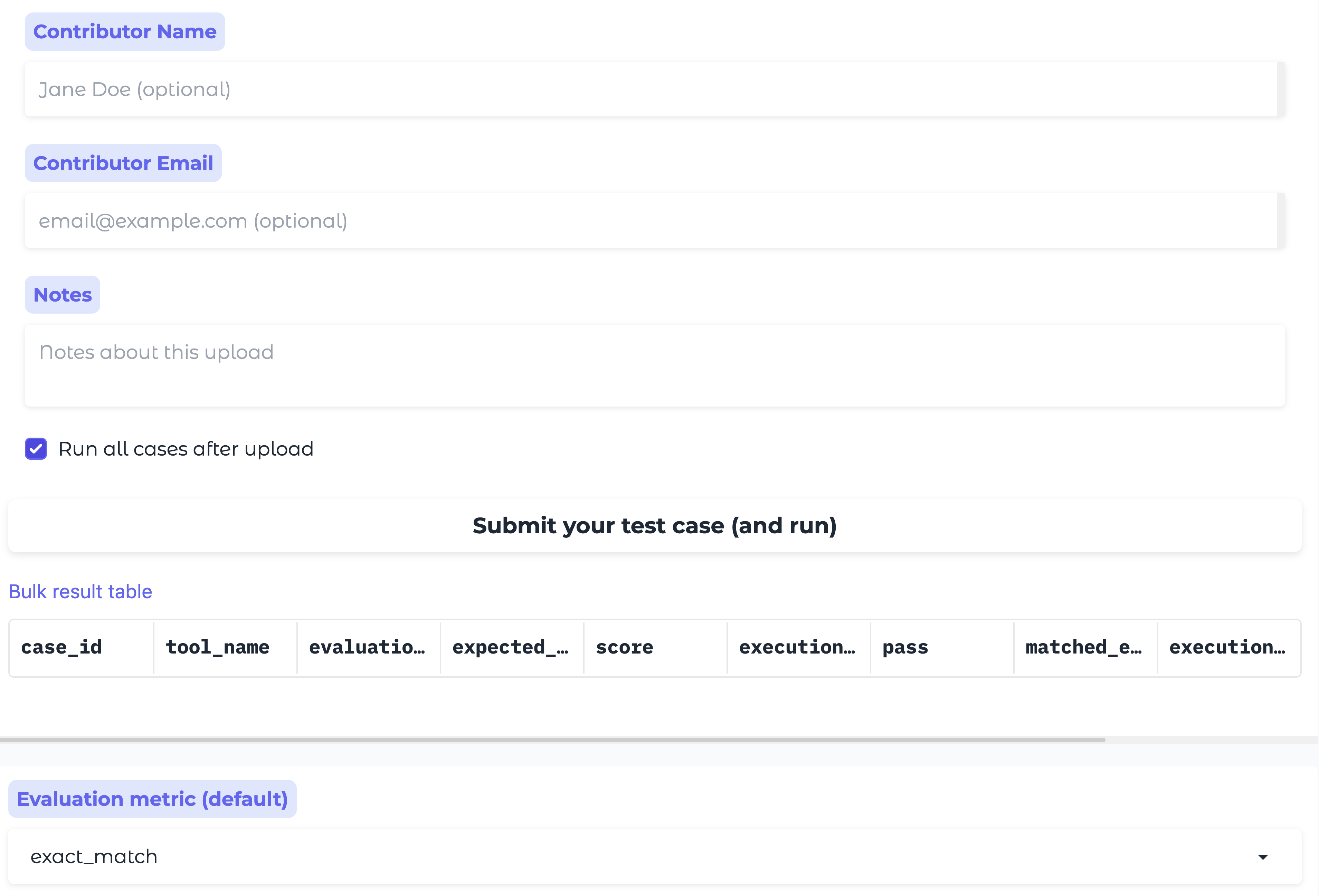}
    \caption{Test-case authoring/execution UI: run tools on submitted cases and store outputs in a bulk-results table with selectable evaluation metrics.}
    \label{fig:author}
\end{figure}

\paragraph{Test Case Contribution}
As described in Section~\ref{sec:demonstrationinterface}, the web interface allows users to contribute test cases for tool evaluation. Submitted cases are reviewed by our team before being added to the shared \textsc{OpenTools} test suite. Figure~\ref{fig:instruct} shows the submission instructions and supported formats. Figure~\ref{fig:author} shows the authoring interface, where users can run tools on submitted cases and store outputs in a bulk-results table. The evaluation uses metrics specified in the submission or selected from a predefined list on the site.

\begin{table*}[!htbp]
\centering
\small
\scriptsize 
\setlength{\tabcolsep}{3pt} 
\renewcommand{\arraystretch}{0.95} 
\begin{adjustbox}{max width=\textwidth, max totalheight=0.85\textheight, keepaspectratio}
\begin{tabular}{l|cc|cc|cc}
\toprule
\textbf{Agent} & \multicolumn{2}{c}{ReAcT} & \multicolumn{2}{c}{OctoTools} & \multicolumn{2}{c}{MultiAgent} \\ 
\midrule
\textbf{Datasets} & \textbf{OctoTools-T} & \textbf{\textsc{OpenTools}-T} & \textbf{OctoTools-T} & \textbf{\textsc{OpenTools}-T} &  \textbf{OctoTools-T} & \textbf{\textsc{OpenTools}-T} \\
\midrule
\multicolumn{7}{c}{\textit{VQA / Puzzle}} \\
\midrule
AlgoPuzzleVQA  & $39.66 {\scriptstyle \pm 2.39}$ / $71.66 {\scriptstyle \pm 0.84}$ & $77.11 {\scriptstyle \pm 0.95}$ / $85.22 {\scriptstyle \pm 1.64}$ & $38.56 {\scriptstyle \pm 2.04}$ / $55.33 {\scriptstyle \pm 2.04}$ & $73.00 {\scriptstyle \pm 0.72}$ /  $86.00 {\scriptstyle \pm 1.54}$ & $70.25 {\scriptstyle \pm 3.03}$ & $75.67 {\scriptstyle \pm 2.02}$  \\
Hallusion-VQA  & $65.56 {\scriptstyle \pm 2.53}$ / $67.50 {\scriptstyle \pm  0.87}$ & $68.00 {\scriptstyle \pm 1.64}$ / $70.33 {\scriptstyle \pm  0.15}$ & $62.78 {\scriptstyle \pm 1.36}$ / $69.00 {\scriptstyle \pm 1.96}$ & $66.00  {\scriptstyle \pm 1.18}$  / $75.33  {\scriptstyle \pm 2.10}$ & $66.55 {\scriptstyle \pm 1.83}$ & $67.33 {\scriptstyle \pm0.82}$  \\
PuzzleVQA     & $36.89 {\scriptstyle \pm 1.75}$ / $72.01 {\scriptstyle \pm 1.93}$ & $41.03 {\scriptstyle \pm 1.28}$ / $73.79 {\scriptstyle \pm 1.85}$ & $48.05 {\scriptstyle \pm 1.06}$ / $70.68 {\scriptstyle \pm 2.57}$ & $46.83  {\scriptstyle \pm 1.75}$ / $73.91 {\scriptstyle \pm 1.13}$ & $71.56 {\scriptstyle \pm 2.97}$ & $73.45 {\scriptstyle \pm 1.93}$ \\
VQA 2.0      & $79.96 {\scriptstyle \pm 1.82}$ / $81.33 {\scriptstyle \pm 0.99}$ & $89.64 {\scriptstyle \pm 0.68}$ / $90.33 {\scriptstyle \pm 1.39}$ & $59.11 {\scriptstyle \pm 1.64}$ / $84.00 {\scriptstyle \pm  1.64}$ & $82.22  {\scriptstyle \pm 1.13}$ / $86.66  {\scriptstyle \pm 1.27}$ & $81.25 {\scriptstyle \pm 1.48}$ & $84.67 {\scriptstyle \pm 1.19}$  \\
\addlinespace[2pt]
\midrule
\multicolumn{7}{c}{\textit{Math / Reasoning}} \\
\midrule
Game of 24   & $76.66 {\scriptstyle \pm 1.32}$ / $100.00{\scriptstyle \pm 0.00}$  & $98.67 {\scriptstyle \pm 0.98}$ / $100.00{\scriptstyle \pm 0.00}$  & $40.89 {\scriptstyle \pm 1.54}$ / $98.33 {\scriptstyle \pm 1.32}$ & $91.22 {\scriptstyle \pm 0.68}$ / $100.00 {\scriptstyle \pm 0.00}$ & $100.00 {\scriptstyle \pm 0.84}$ & $100.00 {\scriptstyle \pm 0.00}$   \\
Omni-MATH    & $27.50 {\scriptstyle \pm 0.37}$ / $74.67 {\scriptstyle \pm 0.65}$ & $33.00 {\scriptstyle \pm 2.05}$  / $76.78 {\scriptstyle \pm 0.41}$ & $30.55 {\scriptstyle \pm 1.59}$ / $71.57 {\scriptstyle \pm 1.87}$ & $29.33 {\scriptstyle \pm 1.15}$ / $73.24 {\scriptstyle \pm 2.56}$ & $70.13{\scriptstyle \pm 1.54}$  & $71.91 {\scriptstyle \pm 1.45}$  \\
CLEVR-Math   & $38.66 {\scriptstyle \pm 1.56}$ / $85.00 {\scriptstyle \pm 1.76}$  & $48.77 {\scriptstyle \pm 1.50}$ / $87.22 {\scriptstyle \pm 1.22}$ & $47.44 {\scriptstyle \pm 1.25}$ / $72.00 {\scriptstyle \pm 1.87}$ & $48.03 {\scriptstyle \pm  1.44}$ / $78.20 {\scriptstyle \pm  1.86}$  & $84.62 {\scriptstyle \pm 1.73}$  & $89.67 {\scriptstyle \pm 2.87}$ \\
MathVista     & $52.00 {\scriptstyle \pm 3.58}$ / $75.00 {\scriptstyle \pm 2.87}$ & $55.67 {\scriptstyle \pm 2.18}$ / $77.33 {\scriptstyle \pm 1.49}$ & $52.84 {\scriptstyle \pm 1.91}$ / $71.00 {\scriptstyle \pm 1.74}$ & $53.56 {\scriptstyle \pm 0.41}$ / $73.67 {\scriptstyle \pm 3.56}$ & $70.23{\scriptstyle \pm 3.51}$ & $71.00 {\scriptstyle \pm 2.34}$\\
\addlinespace[2pt]

\midrule
\multicolumn{7}{c}{\textit{Knowledge / QA}} \\
\midrule
GPQA        & $35.67 {\scriptstyle \pm 2.69}$ / $70.33 {\scriptstyle \pm 3.37}$ & $39.22 {\scriptstyle \pm  4.79}$ / $75.33 {\scriptstyle \pm 0.56}$ & $37.17 {\scriptstyle \pm 2.86}$ / $72.00 {\scriptstyle \pm 3.38}$ & $38.01 {\scriptstyle \pm 0.68}$ / $73.00 {\scriptstyle \pm 1.86}$ & $69.38 {\scriptstyle \pm 3.64}$ & $73.67 {\scriptstyle \pm0.79}$ \\
MMLU-Pro     & $54.33 {\scriptstyle \pm 3.46}$ / $74.67 {\scriptstyle \pm 2.56}$ & $56.69 {\scriptstyle \pm 0.68}$ / $80.00 {\scriptstyle \pm 0.97}$ & $40.67 {\scriptstyle \pm 0.47}$ / $72.33 {\scriptstyle \pm 1.28}$ & $57.33{\scriptstyle \pm 1.02}$ / $74.67 {\scriptstyle \pm 0.96}$ & $73.38 {\scriptstyle \pm 2.37}$ & $78.00 {\scriptstyle \pm 1.34}$ \\
SciFIBench    & $62.00 {\scriptstyle \pm 0.85}$ / $82.00{\scriptstyle \pm 1.07}$ & $64.22 {\scriptstyle \pm 0.87}$ / $84.33 {\scriptstyle \pm 1.18}$ & $65.78 {\scriptstyle \pm 0.56}$ / $84.00 {\scriptstyle \pm 0.56}$ & $63.56{\scriptstyle \pm 0.95}$ / $83.67 {\scriptstyle \pm 0.63}$ & $65.67 {\scriptstyle \pm 0.84}$ & $69.67 {\scriptstyle \pm 1.28}$ \\
\addlinespace[2pt]

\midrule
\multicolumn{7}{l}{\textit{Medical / Multimodal}} \\
\midrule
MedQA       & $76.67 {\scriptstyle \pm 1.09}$ / $92.00 {\scriptstyle \pm 1.27}$ & $79.33 {\scriptstyle \pm  1.73}$ / $94.89 {\scriptstyle \pm 0.15}$ & $77.00 {\scriptstyle \pm 0.40}$ / $92.67 {\scriptstyle \pm 0.43}$ & $76.44 {\scriptstyle \pm 0.47}$ /  $94.33 {\scriptstyle \pm 1.68}$ & $90.11 {\scriptstyle \pm1.25}$ & $93.00 {\scriptstyle \pm 0.23}$  \\
PathVQA     & $31.33 {\scriptstyle \pm 3.63}$ / $45.33 {\scriptstyle \pm 0.68}$ & $39.45 {\scriptstyle \pm  1.36}$ / $47.89 {\scriptstyle \pm 0.81}$ & $35.50 {\scriptstyle \pm 0.81}$ / $42.00 {\scriptstyle \pm 1.37}$ & $36.66 {\scriptstyle \pm 1.02}$ / $43.50 {\scriptstyle \pm 0.36}$ & $44.83 {\scriptstyle \pm 1.18}$ & $46.33 {\scriptstyle \pm 0.49}$ \\
SLAKE       & $48.00 {\scriptstyle \pm 0.25}$ / $70.00 {\scriptstyle \pm 1.35}$ & $54.00 {\scriptstyle \pm  0.47}$ / $71.78 {\scriptstyle \pm 0.83}$ & $46.50 {\scriptstyle \pm 1.77}$ / $65.66 {\scriptstyle \pm 2.37}$ & $53   .33 {\scriptstyle \pm 0.26}$ / $67.67 {\scriptstyle \pm 0.39}$ & $63.21 {\scriptstyle \pm 2.55}$ & $66.67 {\scriptstyle \pm 1.65}$  \\
\addlinespace[2pt]
\midrule
\multicolumn{7}{l}{\textit{General / Agent}} \\
\midrule
GAIA-Text   & $14.17 {\scriptstyle \pm 2.11}$ / $51.97 {\scriptstyle \pm 3.48}$& $28.5 {\scriptstyle \pm  2.15}$ / $55.91 {\scriptstyle \pm 1.61}$ & $4.72 {\scriptstyle \pm 0.64}$ / $46.03 {\scriptstyle \pm 2.53}$ & $24.41 {\scriptstyle \pm 1.56}$ / $53.54 {\scriptstyle \pm 1.75}$& $53.92 {\scriptstyle \pm 3.37}$ & $66.14 {\scriptstyle \pm 2.37}$ \\
\midrule

\rowcolor{gray!15}
\textbf{Average (\%)}  & 49.27 / 74.23 & \textbf{58.22 / 78.08 }& 45.84 / 71.1 &\textbf{ 56.00    / 75.83} & 71.67 & \textbf{75.15} \\
\bottomrule
\\
\end{tabular}
\end{adjustbox}
\caption{Per-benchmark results for 15 benchmarks with GPT-4o-mini and GPT-5-mini as base models. We report mean $\pm$ std over 3 runs; each entry is formatted as GPT-4o-mini / GPT-5-mini. MultiAgent reports a single
gpt-5-mini score. OpenTools-T improves the overall average across all frameworks, demonstrating that higher intrinsic tool quality can translate into better end-to-end reliability. Higher means better.}
\label{tab:per_benchmark_model}

\end{table*}


\begin{figure}[h]
    \centering
    \includegraphics[width=1.8\linewidth]{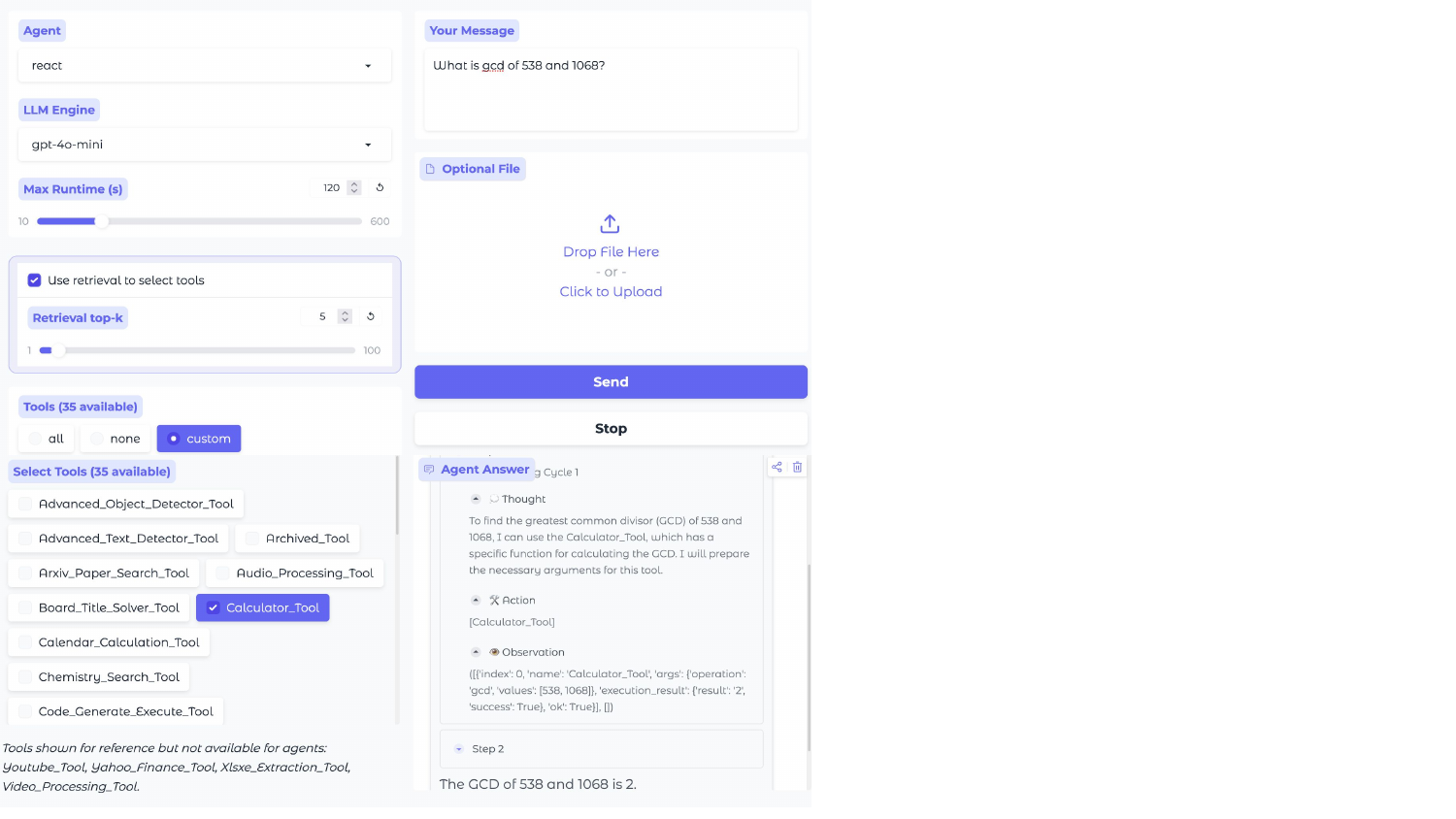}
    \caption{Agent Playground UI: select an agent and base LLM, optionally filter tools via retrieval, and inspect execution logs, responses}
    \label{fig:agent}
\end{figure}
\vspace{-1.5mm}
\paragraph{Agent Workflow}
Figure~\ref{fig:agent} shows the Agent Playground in our web demo, where users select a predefined agent and a base LLM. Users can optionally enable and configure tool retrieval to filter tools before exposing them to the agent. The interface displays agent responses and detailed execution logs in the Agent Answer boxes, and users may also provide feedback on the workflow.
\begin{table}
\centering
\small
\setlength{\tabcolsep}{4.2pt}
\renewcommand{\arraystretch}{1}
\begin{tabular}{l c c}
\toprule
 & \textbf{ZeroShot}
 & \textbf{CoT}\\
\midrule
VQA/Puzzle & 45.38 / \textbf{74.36}&\textbf{46.95} / 74.20 \\
Math/Reasoning &  \textbf{43.77} / \textbf{83.58} & 41.38 / 76.16\\
Scientific   &    46.54 / 80.20 & \textbf{48.89 / 80.74}\\
Medical     &    43.56 / \textbf{69.56} & \textbf{45.33} / 69.33\\
Agent      &     \textbf{ 5.51} / \textbf{21.41} & 4.46 / 20.01\\
\midrule
Average    &     42.16 / \textbf{73.49}  &  \textbf{42.70} / 71.45\\
\bottomrule
\end{tabular}
\caption{Task-group performance for Prompting without tools between ZeroShot and Chain of Thought (CoT). Scores are reported as gpt-4o-mini / gpt-5-mini.}
\label{tab:appendix_zs_cot}
\end{table}
\vspace{-1mm}
\paragraph{Comparison between Prompting Approaches} We compare two tool-free baselines: Zero-shot prompting and Chain-of-Thought (CoT) prompting. CoT elicits explicit step-by-step reasoning using the template in Table~\ref{tab:appendix_zs_cot}. Overall, the two approaches perform similarly; notably, the stronger model (gpt-5-mini) performs better in zero-shot than in CoT (73.49 vs.\ 71.45), suggesting improved intrinsic reasoning without additional instructions.
\vspace{-1.5mm}

\subsection{Experiment Results}
\label{sec:experiment_result}
\paragraph{Comparison Between Toolboxes.}
Table~\ref{tab:per_benchmark_model} reports per-benchmark mean $\pm$ std.\ over three trials for each agent framework under two tool settings, OctoTools-T and \textsc{OpenTools-T}. Consistent with the main results, swapping OctoTools-T for \textsc{OpenTools-T} improves performance across frameworks and domains, indicating that higher tool reliability and coverage translate into better end-to-end agent outcomes. The gains are most pronounced on the \textbf{Agent} benchmark, where robust external actions are required, and they are especially important for the weaker base model (gpt-4o-mini); for the stronger model (gpt-5-mini), improvements remain consistent, though smaller on knowledge/reasoning-heavy tasks.


\end{document}